\documentclass{llncs}
\usepackage{graphicx}
\usepackage{subfigure}
\usepackage{algorithm,algorithmic,eqparbox}
\usepackage{amsmath,amssymb,epsfig}
\usepackage{xcolor}

\begin{document}
\title{Real-Time Predictive Modeling and Robust Avoidance of Pedestrians with Uncertain, Changing Intentions}
\author{Sarah Ferguson\inst{} \and Brandon Luders\inst{} 
\and Robert C. Grande\inst{} \and Jonathan P. How\inst{}}
\institute{Aerospace Controls Laboratory\\
Massachusetts Institute of Technology\\
Cambridge MA 02139}

\maketitle              

\begin{abstract}
To plan safe trajectories in urban environments, autonomous vehicles must be able to quickly assess the future intentions of dynamic agents. Pedestrians are particularly challenging to model, as their motion patterns are often uncertain and/or unknown \emph{a priori}. This paper presents a novel changepoint detection and clustering algorithm that, when coupled with offline unsupervised learning of a Gaussian process mixture model (DPGP), enables quick detection of changes in intent and online learning of motion patterns not seen in prior training data. The resulting long-term movement predictions demonstrate improved accuracy relative to offline learning alone, in terms of both intent and trajectory prediction. By embedding these predictions within a chance-constrained motion planner, trajectories which are probabilistically safe to pedestrian motions can be identified in real-time. Hardware experiments demonstrate that this approach can accurately 
predict pedestrian motion patterns from onboard sensor/perception data and facilitate robust navigation within a dynamic environment.
\end{abstract}

\section{Introduction} \label{sec:intro} 

Autonomous vehicles operating in urban environments must be able to quickly assess the future behavior of nearby agents in order to plan safe trajectories. A major challenge in navigating such environments is the limited ability to accurately anticipate the intents of dynamic agents, as their internal state and desires are not directly observable. Dynamic agents exhibit uncertain in both their intent and the trajectory motion pattern associated with each intent. Pedestrians present particular technical challenges in the generation of long-term predictions, specifically (i) the demonstration of many unique behaviors, which may not have been previously observed; and (ii) instantaneous changes in motion behavior following changes in intent.

This paper addresses these challenges by proposing a novel framework for long-term predictions of pedestrian motion, capable of learning new motion patterns and quickly responding to changes in intent as they are observed online. Such a predictive model is necessary because naively modeling the uncertainty in dynamic environments can lead to rapidly increasing uncertainty, which is typically prohibitive for safe navigation~\cite{trautman2010unfreezing}. The preferred approach in the literature, also used here, assumes that factors influencing pedestrian motion (such as internal state and intent) are reflected by their trajectories. Therefore, various data-driven approaches learn typical motion patterns from observed training trajectories to enable predictions of future state.

The most common approaches are based on the Markov property. This set of approaches includes hidden Markov models, in which the hidden state is pedestrian intent~\cite{bennewitz2005learning,kelley2008understanding,zhu1991hidden}; growing hidden Markov models to allow for online learning~\cite{vasquez2009incremental}; and partially observable Markov decision processes to choose actions based on a distribution over pedestrian intents~\cite{bandyopadhyayintention}. Because the future state prediction is dependent only on the current state, these approaches are quick to react to changes in intent. However, for relatively infrequent changes in intent, the Markov assumption can be overly restrictive, as it prevents these algorithms from becoming more certain of pedestrian intent with additional observations. 

Gaussian process (GP) approaches have been demonstrated to be well-suited for modeling pedestrian motion patterns, as they perform well with noisy observations and have closed-form predictive uncertainty~\cite{ellis2009modelling,trautmanrobot,RasmussenWilliams2005}. Additionally, recent work using GP mixture models enables predictions that account for both intent and trajectory uncertainty~\cite{Aoude13_AURO}. Both sets of approaches use the entire observed trajectory in the prediction of future state, such that certainty in demonstrated intent tends to converge over time.  Therefore, when changes in intent occur, these approaches are much slower to detect a change than Markov-based approaches. Additionally, existing GP classification approaches are too slow for online learning of previously unobserved behavior patterns.

This paper proposes a novel changepoint detection and clustering algorithm which retains the trajectory prediction accuracy of existing GP approaches while expanding their capabilities. Coupled with offline unsupervised learning of a Gaussian process mixture model (DPGP)~\cite{Joseph11_AR}, this approach enables quick detection of changes in intent and online learning of motion patterns not seen in prior training data. The resulting long-term movement predictions demonstrate improved accuracy relative to offline learning alone in both intent and trajectory prediction.
These predictions can also be used within a chance-constrained motion planner~\cite{Luders10_GNC} to identify probabilistically safe trajectories in real-time. In experimental results, the proposed algorithm is used to predict the motion of multiple dynamic agents detected from a variety of onboard and external sensors, enabling an autonomous rover to robustly navigate dynamic environments.

\section{Preliminaries} \label{sec:prelim}

\subsection{Motion Patterns and Modeling} \label{sec:motion}
A pedestrian trajectory is represented as a set of observed $xy$-locations $(x^i_1,y^i_1),$ $(x^i_n,y^i_n),$ ..., $(x^i_{L^i},y^i_{L^i})$, where $L^i$ is the total length of the trajectory $t^i$ of pedestrian $i$.  Trajectories are not necessarily of the same length, and the time steps between each observation may be irregular.  A motion pattern is defined as a mapping from $xy$-locations $(x^i,y^i)$ to a distribution over trajectory derivatives $\left( \frac{\Delta x^i}{\Delta t}, \frac{\Delta y^i}{\Delta t}\right)$, resulting in a velocity flow-field in $x-y$ space.  This approach is therefore independent of the lengths and discretization of the trajectories.  

This work uses Gaussian processes (GP) for its motion pattern models.  Although GPs have a significant mathematical and computational cost, they generalize well to regions of sparse data while avoiding the problem of over fitting in regions of dense data.  GP models are extremely robust to unaligned, noisy measurements and are well-suited for modeling the continuous paths underlying potentially non-uniform time-series samples of pedestrian locations~\cite{RasmussenWilliams2005}.  

The GP serves as a non-parametric form of interpolation between the discrete trajectory measurements.  Specifically, given an observed $(x,y)$ location, the GP predicts the trajectory derivatives at that location.  The standard squared exponential covariance function describes the correlation between trajectory derivatives at two points $(x,y)$ and $(x',y')$. The mean trajectory derivative functions $E[\frac{\Delta x^i}{\Delta t}, \frac{\Delta y^i}{\Delta t}]=\mu_x(x,y)$ and $E[\frac{\Delta y^i}{\Delta t}, \frac{\Delta y^i}{\Delta t}]=\mu_y(x,y)$ are implicitly initialized to zero for all $xy$ locations.

The resulting motion model is defined as a finite mixture of the $M$ learned motion patterns weighted by their prior probabilities.  The finite mixture model probability of the $i$th observed trajectory $t^i$ is 
\begin{eqnarray}
 \label{eq:motionmodel} p(t^i)=\sum_{j=1}^M p(b_j) p(t^i | b_j),
\end{eqnarray}
where $b_j$ is the $j$th motion pattern and $p(b_j)$ is its prior probability. The prior $p(b_j)$ is initialized to be proportional to the number of trajectories in motion model $j$, but is updated after each prediction with the previous posterior probability which incorporates the GP likelihood. The number of motion patterns $M$ is learned offline via an automated clustering process and incremented as new behavior patterns are identified online. This motion model has been previously presented in ~\cite{Aoude13_AURO,Joseph11_AR}.

Future pedestrian trajectories are predicted for each motion pattern using the approach of ~\cite{Girard03_NIPS,Deisenroth2009a}.  This approach provides a fast, analytic GP approximation specifying possible future pedestrian locations $K$ time steps ahead, while incorporating uncertainty in previous predictions at each time step.

\subsection{Batch Learning of Motion Patterns} \label{sec:dpgp}

It is expected that observed pedestrian trajectories will demonstrate a variety of qualitatively different behaviors. These behavior motion patterns are learned from an input set of unlabeled trajectories by DPGP, a Bayesian nonparametric clustering algorithm that automatically determines the most likely number of clusters without \emph{a priori} information~\cite{Joseph11_AR}. This section reviews the DPGP algorithm, which is used in this work to cluster observed pedestrian trajectories into representative motion patterns in batch.

The DPGP algorithm models motion patterns as Gaussian processes, as described in Sect.~\ref{sec:motion}. A Dirichlet process (DP) mixture model potentially allows for an infinite mixture of motion patterns, where the DP concentration parameter $\alpha$ controls the probability of new cluster formation. A smaller $\alpha$ enforces the expectation that there are a few motion patterns that pedestrians tend to exhibit; therefore, trajectories are more likely to fit existing clusters than to form new ones.

Because exact inference over the space of GPs and DPs is intractable, samples are drawn from the posterior over motion models to train the model.  At each iteration, the DP hyperparameter $\alpha$ is resampled using Gibbs sampling techniques and the GP hyperparameters for the $j$ behavior patterns $\theta_{x,j}^{GP},\theta_{y,j}^{GP}$ are set to their maximum likelihood values using gradient ascent. The probability that trajectory $t^i$ will be assigned to an existing motion pattern is
\begin{eqnarray}
	p(z_i=j|t^i,\alpha,\theta_{x,j}^{GP},\theta_{y,j}^{GP}) \propto l(b_j;t^i)\left(\frac{n_j}{N-1+\alpha}\right),
\end{eqnarray}
and the probability that trajectory $t^i$ will be assigned to a new motion pattern is
\begin{eqnarray}
	p(z_i=M+1|t^i,\alpha) \propto \int l(b_j;t^i) d\theta_{x,j}^{GP} d\theta_{y,j}^{GP} \left(\frac{\alpha}{N-1+\alpha}\right),
\end{eqnarray}
where $l(b_j;t^i)$ is the likelihood that trajectory $t^i$ is assigned to motion pattern $b_j$, $n_j$ is the number of trajectories currently assigned to $b_j$, and $N$ is the total number of trajectories. The likelihoods $l(b_j;t^i)$ can be determined from the GP trajectory derivatives. 

\subsection{Motion Planning} \label{sec:ccrrt}

Motion planning for autonomous vehicles is executed via chance-constrained rapidly-exploring random trees (CC-RRT), which can efficiently identify trajectories with guaranteed minimum bounds on constraint satisfaction probability under internal and/or external uncertainty~\cite{Luders10_GNC}. The primary objective is to plan and execute a motion plan directing the vehicle to reach some goal region, while ensuring the non-convex state constraints $x_t \in \mathcal{X}_t$ are probabilistically satisfied. This is represented via path-wise and time-step-wise chance constraints
\begin{eqnarray}
 \label{eq:ccpathstep} \mathbb{P} \left ( \bigwedge_t x_t \in \mathcal{X}_t \right ) \geq \delta_p, ~~~~~~~~ \mathbb{P} \left ( x_t \in \mathcal{X}_t \right ) \geq \delta_s,~~\forall~t,
\end{eqnarray}
respectively, where $\mathbb{P}(\cdot)$ denotes probability, $\bigwedge$ represents a conjunction over the indexed constraints, and $\delta_s,\delta_p \in [0.5,1]$ are chosen by the user. The feasible state space $\mathcal{X}_t$ consists of a convex environment containing multiple convex, polytopic obstacles to be avoided. It is assumed that the shape and orientation of these obstacles is known, but their placement may be uncertain and/or dynamic.

The CC-RRT algorithm samples a tree of dynamically and probabilistically feasible trajectories through the environment, rooted at the vehicle's current state. All trajectories added to the tree must satisfy (\ref{eq:ccpathstep}), which CC-RRT evaluates by leveraging the trajectory-wise constraint checking of sampling-based algorithms to efficiently compute risk bounds~\cite{Luders10_GNC}. 

In this work, detected pedestrians are modeled as dynamic obstacles, with both intent and trajectory uncertainty as represented by (\ref{eq:motionmodel}). This model provides a likelihood and time-parameterized uncertainty distribution for each behavior of each pedestrian obstacle. The CC-RRT formulation has been recently expanded to guarantee probabilistically robust avoidance of dynamic obstacles with uncertain intentions, making it suitable for robust avoidance of pedestrian models. In particular, if each dynamic pedestrian behavior is treated as a separate obstacle whose risk bound is scaled by its likelihood, then all probabilistic feasibility guarantees are maintained~\cite{Aoude13_AURO}.

\section{Changepoint Detection} \label{sec:nbht}

To effectively anticipate the motion of pedestrians, this paper proposes a framework which can perform online classification of observed trajectories, in addition to learning common pedestrian trajectories from batch data.
Because agile dynamic agents such as pedestrians may exhibit new behaviors or mid-trajectory changes in intent, this problem is framed in the context of changepoint detection.

This work utilizes a variation of the generalized likelihood test (GLR)~\cite{basseville1995detection} to perform changepoint detection. The basic GLR algorithm detects changes by comparing a windowed subset of data to a null hypothesis. If the maximum likelihood statistics of the windowed subset 
differ from the null hypothesis significantly, the algorithm returns that a changepoint has occurred. 

The proposed changepoint detection algorithm is given in Algorithm \ref{alg:NBNP}. At each time step, given Gaussian process $GP_w$, the algorithm creates a new GP ($GP_S$) with the same hyperparameters, but using a windowed data subset $S$ of size $m_S$ (lines 2--4). Although $m_S$ is domain specific, the algorithm is fairly robust to its selection; $m_S \approx 10-20$ works well for most applications.

The algorithm then calculates the joint likelihood of the set having been generated from the current GP model (the null hypothesis $H_0$) and the new $GP_S$ ($H_1$). At each step, the normalized log-likelihood ratio test (LRT) is computed as
\begin{equation}
L(y) = \frac{1}{m_s}(\log P(\mathcal{S} \mid H_1) - \log P(\mathcal{S} \mid H_0)). \label{eq:LRT}
\end{equation}
For a GP, the log likelihood of a subset of points can be evaluated in closed form as
\begin{eqnarray*}
\log P(y \mid x, \Theta) = -\frac{1}{2}(y-\mu(x))^T\Sigma_{xx}^{-1} (y-\mu(x)) - \log |\Sigma_{xx}|^{1/2} + C,
\end{eqnarray*}
where $\mu(x)$ is the mean prediction of the GP and
\begin{eqnarray*}
\Sigma_{xx} = K(x,x) +\omega_n^2I -K(X,x)^T(K(X,X) + \omega_n^2 I)^{-1}K(X,x)
\end{eqnarray*}
is the predictive variance of the GP plus the measurement noise. The first term of the log-likelihood accounts for the deviation of points from the mean, while the second accounts for the relative certainty (variance) in the prediction.

\begin{algorithm}[t]
\caption{Likelihood Ratio Test}
\begin{algorithmic}[1]
\label{alg:NBNP}
\STATE {\bf Input}: Set of points $\mathcal{S}$, Working model $GP_w$
\STATE $l_1 = \log p(\mathcal{S} \mid GP_w)$
\STATE Create new GP $GP_{S}$ from $\mathcal{S}$ \label{alg:new_GP}
\STATE $l_2 = \log p(\mathcal{S} \mid GP_{S})$
\STATE Calculate LRT $L_i(y) = \frac{1}{m_S}\left( l_2 - l_1 \right)$  
\STATE Calculate average of last $m$ LRT: \\ $\qquad L_m =\frac{1}{m}\sum_{j=i-m}^{i}L_j(y)$ \label{alg:avg}
\STATE Calculate average of LRT after changepoint: \\ $\qquad L_{ss} =\frac{1}{i-m-1}\sum_{j=1}^{i-m-1}L_j(y)$ \label{alg:ss}
\STATE $i = i+1$
\RETURN{  $L_m - L_{ss} \geq \eta$ \label{alg:LRT}}
\end{algorithmic}
\end{algorithm}

Algorithm \ref{alg:NBNP} uses the LRT to determine if the maximum likelihood statistics (mean and variance) of $GP_S$ differ significantly from the null hypothesis, indicating that the points in $\mathcal{S}$ are more unlikely to have been generated from the model $GP_w$. In particular, the average over the last $m$ LRT values (line \ref{alg:avg}) is compared to the nominal LRT values seen up until this point (line \ref{alg:ss}). If the difference of these two values exceeds some value $\eta$, the algorithm returns false, indicating that this generating model does not fit the data. The value $\eta$ can be determined based on the probability of false alarms and maximum allowed error~\cite{Grande14_UAI}. 

The LRT algorithm is quite robust in practice, based on the following intuition. If the points in $S$ are anomalous simply because of output noise, then the new GP model created from these points will on average be similar to the current model. Additionally, the joint likelihood given the new model will not be substantially different from that of the current model.
However, if the points are anomalous because they are drawn from a new process, then the resulting GP model will on average be substantially different from the current model, yielding a higher joint likelihood of these points. Lastly, instead of making a decision on a single LRT, the last $m$ LRT's are averaged and compared to the average LRT values seen since the last changepoint. In practice, the LRT may have some offset value due to modeling error. Looking at the difference between the last $m$ values and the average LRT values makes the algorithm robust to this problem.

\section{Changepoint-DPGP} \label{pedmodel}

In order to generate accurate and computationally efficient predictions of pedestrian motion, the proposed Changepoint-DPGP algorithm seeks to identify new behaviors online and detect changes in intent given typical pedestrian behaviors learned from batch data. The key idea behind this algorithm is to perform online classification of a sliding window of trajectory segments, and detect changepoints or new behavior models according to changes in the current classification.

\begin{algorithm}[t]
\caption{Changepoint-DPGP}
\begin{algorithmic}[1]
\label{alg:cpdpgp}
\STATE {\bf Input}: Set of previous behavior models $\mathcal{GP} = \{GP_1, \hdots, GP_N\}$
\WHILE{Input/Output $\langle x_t, y_t \rangle$ available}
	\STATE Add $\langle x_t, y_t \rangle$ to $\mathcal{S}$
	\STATE Call Algorithm~\ref{alg:compare}
	\IF[Change in intent detected]{$\mathcal{M}_{t-1} \cap \mathcal{M}_t = \varnothing$}
		\STATE Reinitialize priors \label{alg:switch_intent}
	\ENDIF
	\IF[New behavior detected]{$\mathcal{M}_t = \varnothing$}
		\STATE Initialize new model $GP_n$
	\ELSE
		\STATE Predict according to Sect.~\ref{sec:motion}
	\ENDIF
	\IF{$\mathcal{M}_t \neq \varnothing$}
		\STATE $\mathcal{M}_t = \mathcal{M}_{t-1} \cap \mathcal{M}_t$
	\ENDIF
\ENDWHILE
\IF{$GP_n$ is initialized}
	\STATE Add $\langle x_{0:T}, y_{0:T} \rangle$ to $GP_n$
	\STATE Add $GP_n$ to set of current models $\mathcal{GP}$
\ENDIF
\end{algorithmic}
\end{algorithm}

\begin{algorithm}[t]
\caption{Compare to Current Models}
\begin{algorithmic}[1]
\label{alg:compare}
\STATE {\bf Input}: Set of current behavior models $\mathcal{GP} = \{GP_1, \hdots, GP_N\}$
\STATE Initialize representative model set $\mathcal{M}_t$
\FOR{Each $GP_j \in \mathcal{GP}$}
	\STATE Call Algorithm~\ref{alg:NBNP} with inputs $\mathcal{S}$, $GP_j$
	\IF{Algorithm~\ref{alg:NBNP} returns \TRUE}
		\STATE Add $GP_j$ to $\mathcal{M}_t$
	\ENDIF
\ENDFOR
\end{algorithmic}
\end{algorithm}

The Changepoint-DPGP algorithm is detailed in Algorithm~\ref{alg:cpdpgp}. The algorithm begins with an initial set of learned behavior motion models $\mathcal{GP}$, obtained from running the DPGP algorithm on batch training data. As new data points are received, they are added to a sliding window $\mathcal{S}$ of length $m_s$. After creating a new model $GP_S$ from the points in $\mathcal{S}$, the LRT is computed for $GP_S$ and for each model $GP_j$ in the current model set $\mathcal{GP}$. This process determines if the points in $\mathcal{S}$ are statistically similar to those in the model $GP_j$, subject to the predetermined threshold $\eta$. 

In order to detect changepoints, the algorithm maintains the set of models $\mathcal{M}_t$ that the points in $\mathcal{S}$ fit into at each time step, representative of the current classification of those points. Because the behavior patterns may overlap (e.g. the blue/green and red/teal behavior patterns in Fig.~\ref{fig:crosswalk-velodata}), a single classification cannot be guaranteed, necessitating the maintenance of a model set. Changepoints occur when the classification changes, i.e. when the current classification $\mathcal{M}_t$ and previous classification $\mathcal{M}_{t-1}$ share no common models. Additionally, the current classification is reset at each timestep to be the intersection of the current and previous classification sets, assuming that the current classification is not empty.

To illustrate this method, consider a pedestrian crossing a crosswalk by following the green behavior pattern in Fig.~\ref{fig:crosswalk-velodata}. Until the pedestrian reaches the crosswalk, $\mathcal{M}_t=\{B,G\}$. Once the pedestrian enters the crosswalk, their classification becomes $\mathcal{M}_t=\{G\}$. A changepoint should not be detected at this stage, as the pedestrian is committing to the green behavior rather than changing their intent. However, if the pedestrian switched to the teal behavior after entering the crosswalk, this would represent a change in intent. The classification for three successive timesteps would become $\mathcal{M}_{t-2}=\{G\}$, $\mathcal{M}_{t-1}=\{G,T\}$, $\mathcal{M}_t=\{T\}$ and no changepoint would be detected if ${\mathcal T}_t$ was not reset.

The predictive component of this algorithm is decoupled from classification. In general, the future state distribution is computed as described in Sect.~\ref{sec:prelim}. However, if at any point $\mathcal{M}_t$ is empty, this indicates that the current model set $\mathcal{GP}$ is not representative for the points in $\mathcal{S}$, so a new behavior must be created. The algorithm waits until the entire new trajectory has been observed to create the new behavior pattern, generating predictions according to a simple velocity propagation model until the model set becomes representative. In practice, any reasonable predictive model can be used at this stage, as no information on the anomalous agent's current behavior.

If the training data contains trajectories with changes in intent, DPGP will learn unique behavior patterns for each trajectory containing a changepoint, as the entire trajectory is considered for classification. To obtain a representative set of behavior patterns, the Changepoint-DPGP algorithm can be used offline to reclassify these trajectories by segmenting them at the changepoint. To do so, Algorithm~\ref{alg:cpdpgp} is first called with $\mathcal{GP}$ containing those behavior patterns with more than $k_{min}$ trajectories and data $\langle x_t, y_t \rangle$ from trajectories in the remaining behavior patterns not in $\mathcal{GP}$. At line~\ref{alg:switch_intent} and at the end of Algorithm \ref{alg:cpdpgp}, the trajectory segment seen since the last changepoint is classified into the most likely behavior pattern. The intuition behind this modification is that changes in intent are agent-specific; therefore, behavior patterns containing these trajectories are not representative of global behaviors 
caused by the environment.

As the Changepoint-DPGP algorithm is called for each dynamic agent, with learned motion models in $\mathcal{GP}$ possibly shared among all agents, Algorithms \ref{alg:cpdpgp}-\ref{alg:compare} can be parallelized to speed up computation if desired.

\section{Results} \label{sec:experiments}
This section presents empirical results which evaluate Changepoint-DPGP on real-world problem domains of varying complexity. The prediction results demonstrate that prior observations of pedestrian motion can be used to learn accurate behavior models. These models are applied to real-time observations to make accurate, long-term predictions of complex motion behavior, beyond what could be predicted from the observations themselves (e.g., bearing and speed). The planner is then demonstrated to select safe paths which are risk-aware with respect to possible pedestrian intentions, their likelihood, and their risk of interaction with the host vehicle. The final version of this paper will include more comprehensive results and analysis.

\subsection{Pedestrian Crosswalk}
\begin{figure}[t]
 \centering
 \subfigure[Environment for crosswalk experiments. Rover starts in foreground, while pedestrian follows one of four possible behaviors (red). Velodyne location is marked with green arrow.]{
 \includegraphics[trim=30mm 0mm 30mm 0mm, clip, width=0.52\linewidth]{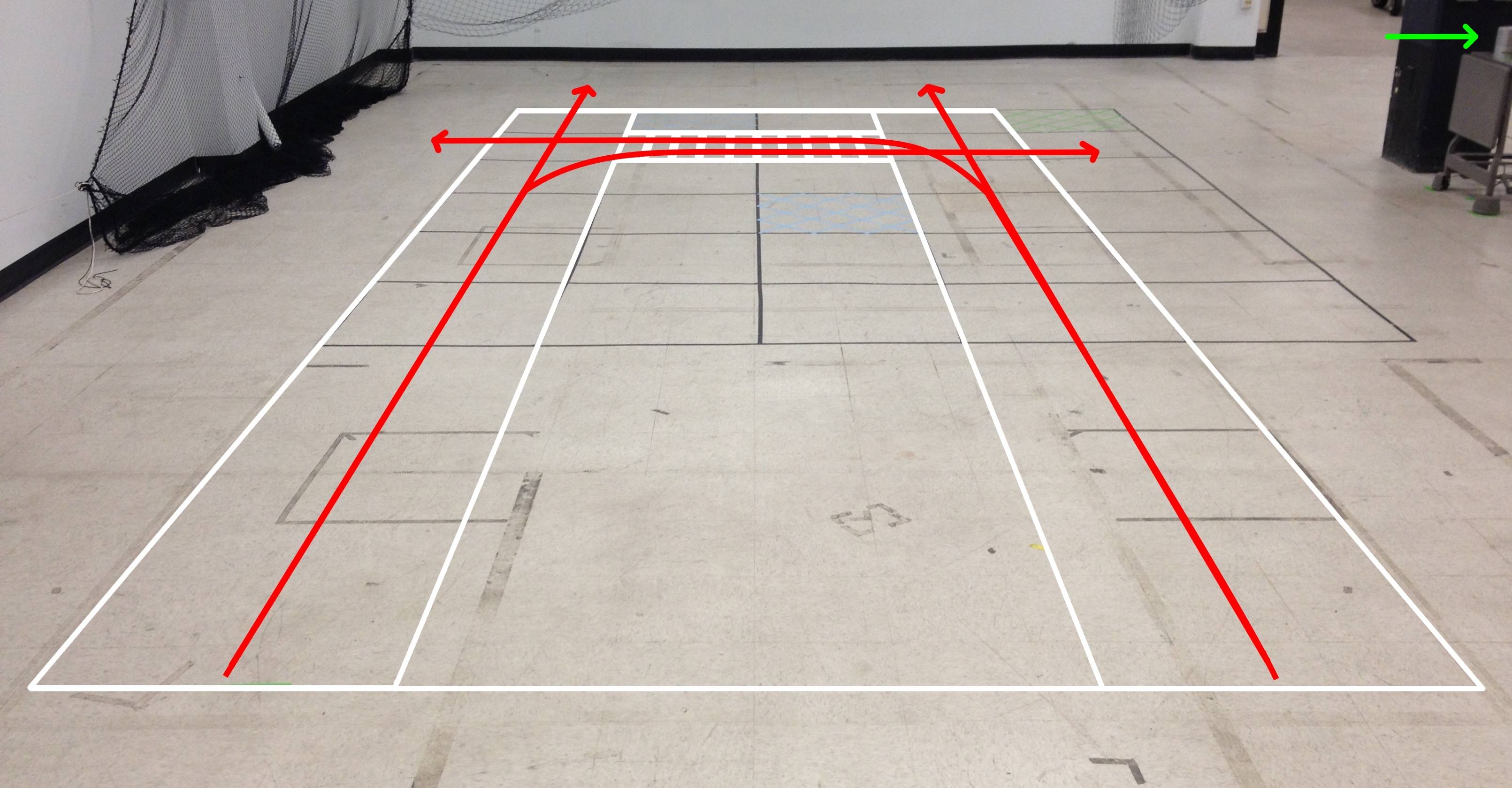}
 \label{fig:crosswalk-bldg41}
 } \hspace{1mm}
 \subfigure[Training pedestrian trajectories collected by Velodyne lidar and resulting DPGP velocity flow fields for each behavior (separated by color).]{
  \includegraphics[width=0.42\linewidth,trim = 0 20 0 20,clip]{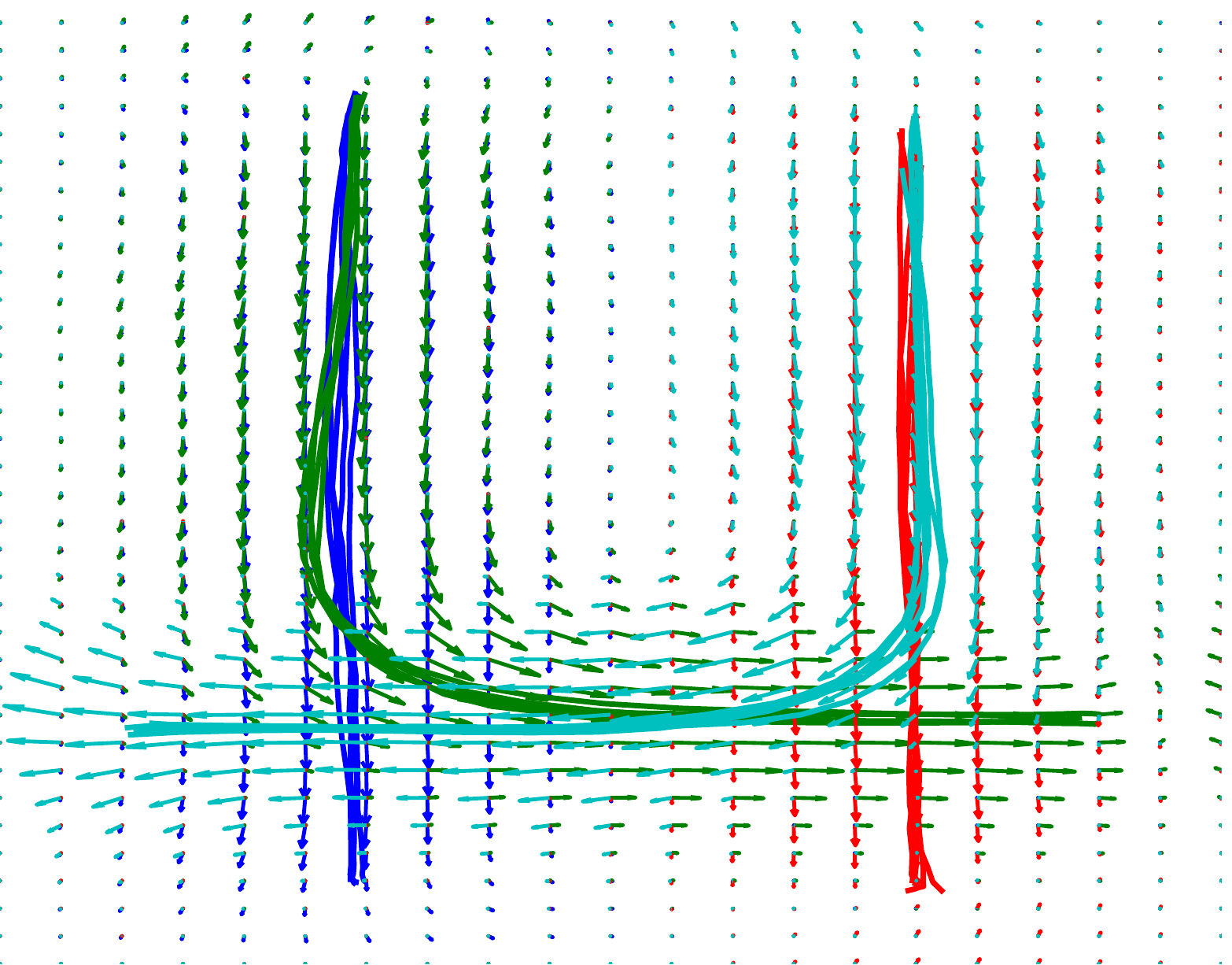}
 \label{fig:crosswalk-velodata}
 }
 \caption[]{Environment setup and pedestrian data for crosswalk experiments.}
 \label{fig:crosswalk-setup}
 \vspace{-2mm}
\end{figure}

Consider the scenario in Fig.~\ref{fig:crosswalk-setup}, in which an autonomous rover travels along a street flanked by two sidewalks and must safely pass through a pedestrian crosswalk. Pedestrians have four possible behaviors (red) corresponding to which sidewalk they are traversing, and whether they choose to use the crosswalk.

A Pioneer 3-AT rover is used as the autonomous vehicle in this and subsequent experiments. Its payload includes a SICK LMS-291 lidar for onboard pedestrian detection and an Intel Core i5 laptop with 6GB RAM for computation. The online perception, planning, and control algorithms described in this paper are executed on this laptop via the Robotic Operating System (ROS)~\cite{quigley2009ros}. Dynamic obstacle detections and autonomous vehicle state are fed to a real-time, multi-threaded Java application, which executes the chance-constrained planner (Sect. \ref{sec:ccrrt}) to generate safe paths. A pure pursuit controller~\cite{Park07_JGCD} generates acceleration commands to follow this set of waypoints.  High-fidelity localization is provided for the rover via motion-capture cameras~\cite{How08_CSM}.

Three trajectory prediction algorithms are evaluated in this experiment: changepoint-DPGP, DPGP, and a goal-directed approach using hidden Markov models (HMM)~\cite{Michini13_ICRA}. The hidden states of the HMM are pedestrian goals, learned via Bayesian nonparametric inverse reinforcement learning with an approximation to the action likelihood specifying that pedestrians head directly towards goal locations~\cite{Michini13_ICRA}. This motion model assumes that each pedestrians head directly toward their intended goal at some preferred speed with an uncertainty distribution over heading and velocity, as used by~\cite{helbing1995social,bandyopadhyayintention,ikeda2012modeling} among others.

Unless otherwise noted, all three algorithms were trained on five trajectories from each of the four behavior patterns in Fig. \ref{fig:crosswalk-bldg41}. Each trajectory was collected by taking observations of an actual pedestrian traversing the environment, as observed by a Velodyne HDL-32E lidar at the location marked in green in Fig.~\ref{fig:crosswalk-bldg41}. Pedestrians are identified from the raw Velodyne returns both offline and online using Euclidean clustering~\cite{rusu20113d}. Fig. \ref{fig:crosswalk-velodata} shows the training trajectories used in this experiment.

\begin{figure}[t]
 \centering
 \subfigure[Probability of correct motion pattern]{
 \includegraphics[width=0.475\linewidth]{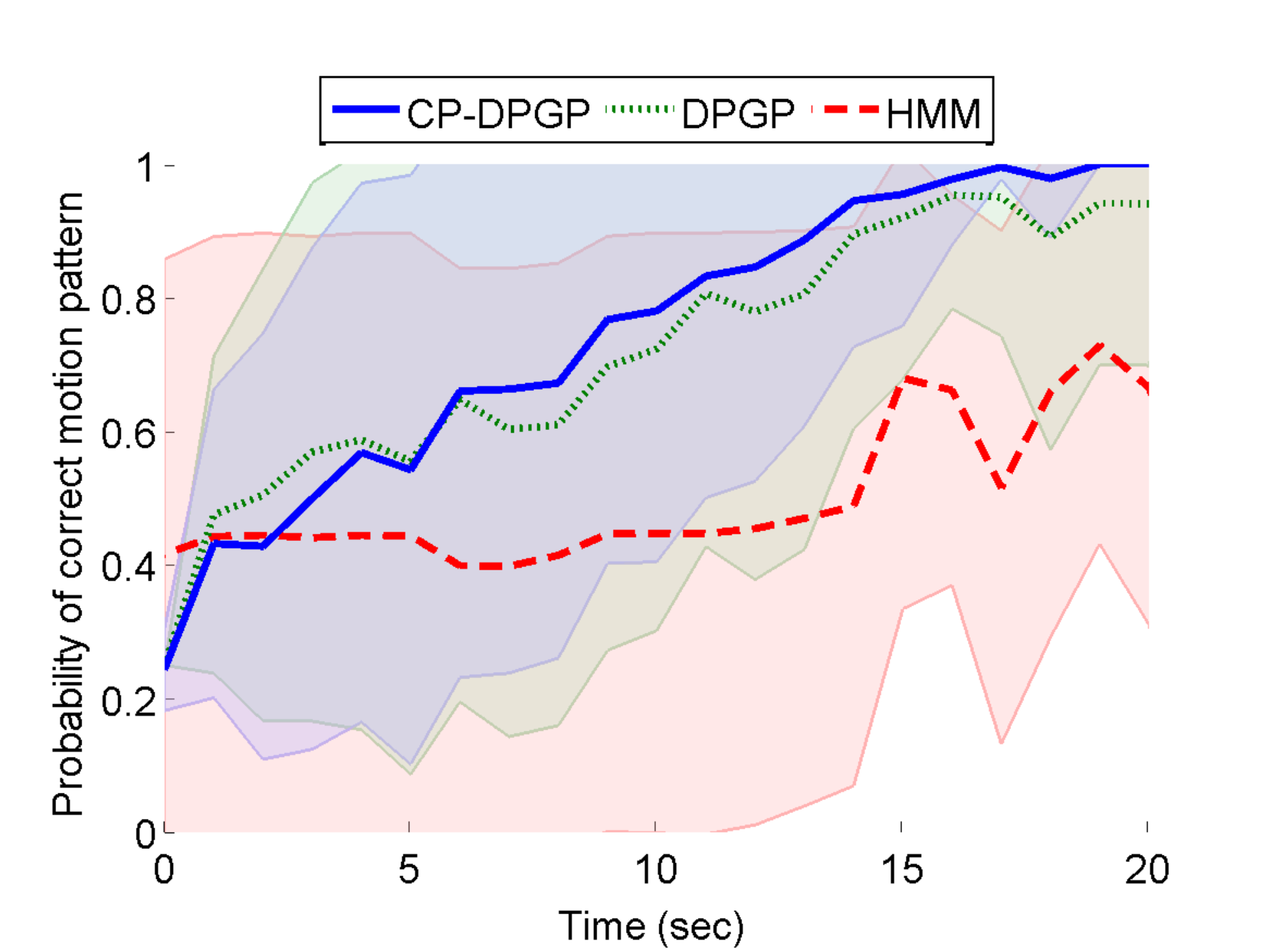}
 \label{fig:baseline_intent}
 }
\subfigure[RMS predictive error]{
  \includegraphics[width=0.475\linewidth]{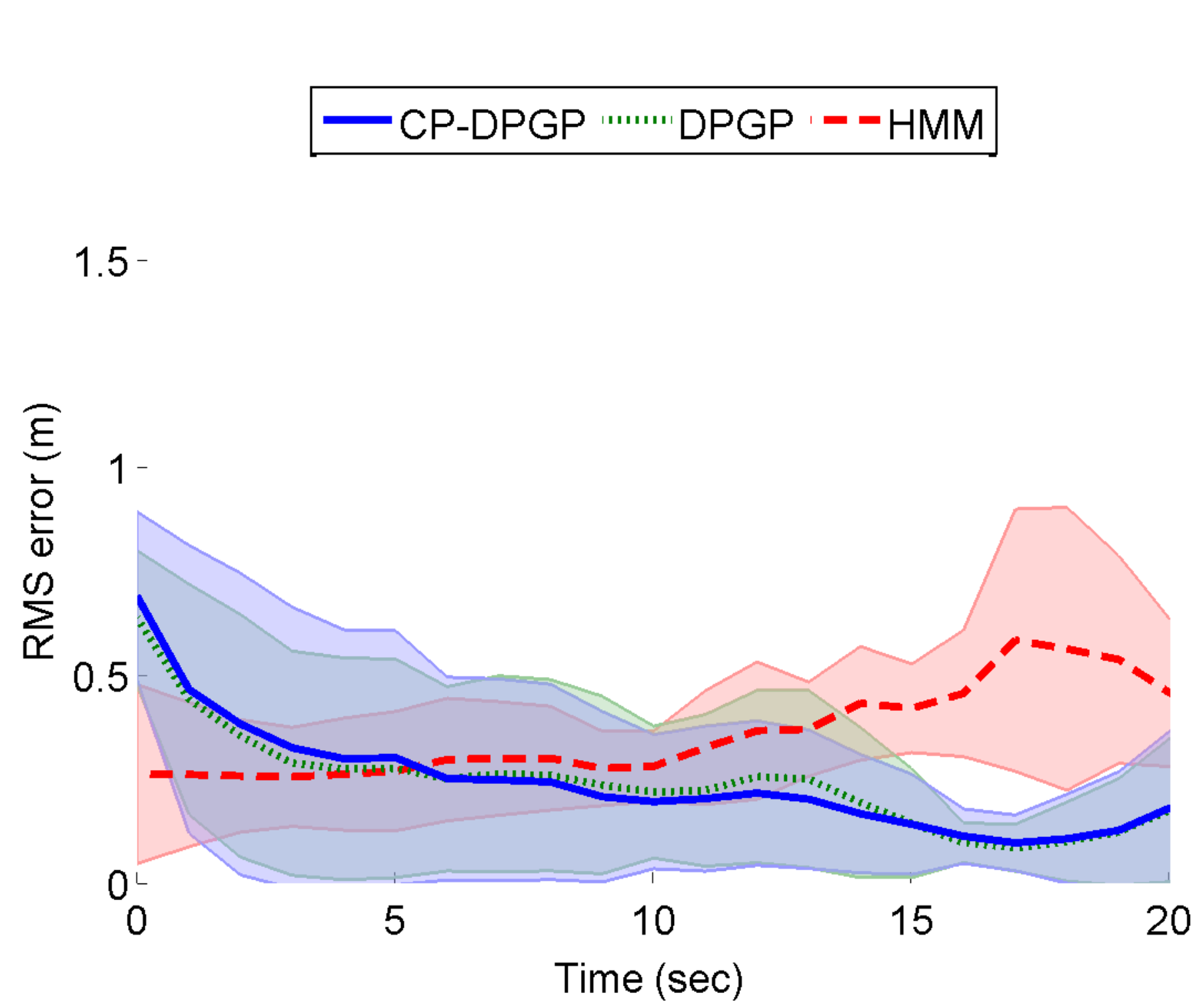}
 \label{fig:baseline_rms}
 }
 \caption[]{Prediction accuracy of each algorithm for the baseline case of the pedestrian crosswalk scenario. All results are averaged over 20 trials as a function of time elapsed, with error bars representing standard deviation.}
 \label{fig:baseline}
  \vspace{-2mm}
\end{figure}

Fig. \ref{fig:baseline} considers the baseline case in which (i) neither the training nor test data exhibits any mid-trajectory changes in intent; and (ii) the test data does not exhibit any behaviors unseen in the training data. In these results, each algorithm is tested on five trajectories from each of the four behavior patterns (Fig. \ref{fig:crosswalk-bldg41}). Fig. \ref{fig:baseline_intent} displays the probability each algorithm has assigned to the correct motion pattern given the observation trajectory, averaged across all 20 trials as a function of time elapsed. This metric measures the ability of each algorithm to identify the correct pedestrian intentions. The likelihoods of each motion pattern serve as the intent prediction for the GP-based approaches, with the prior probability (time = 0) based on the fraction of training trajectories for each motion pattern. The most likely state distribution, calculated via the forward algorithm and Markov chain propagation, describes the predicted 
intent for the HMM approach. Fig. \ref{fig:baseline_rms} displays the root mean square (RMS) error between the true pedestrian position and the mean predicted position (Sect. ~\ref{sec:motion}), averaged across all 20 trials as a function of time elapsed. This metric captures overall prediction accuracy subject to both intent and path uncertainty.

The Markov property prevents the HMM approach from converging to the correct motion pattern, as the observations of current state alone are not sufficient in the case of noisy observations (Fig. \ref{fig:baseline_intent}). As a result, its RMS error tends to increase over time. On the other hand, both GP approaches exhibit convergence in the probability of the correct motion pattern as new observations are made, which improves RMS predictive error as well. The performance of changepoint-DPGP and DPGP is very similar, as is expected in the absence of changepoints and new behaviors.

\begin{figure}[t]
 \centering
 \subfigure[Probability of correct motion pattern]{
 \includegraphics[width=0.475\linewidth]{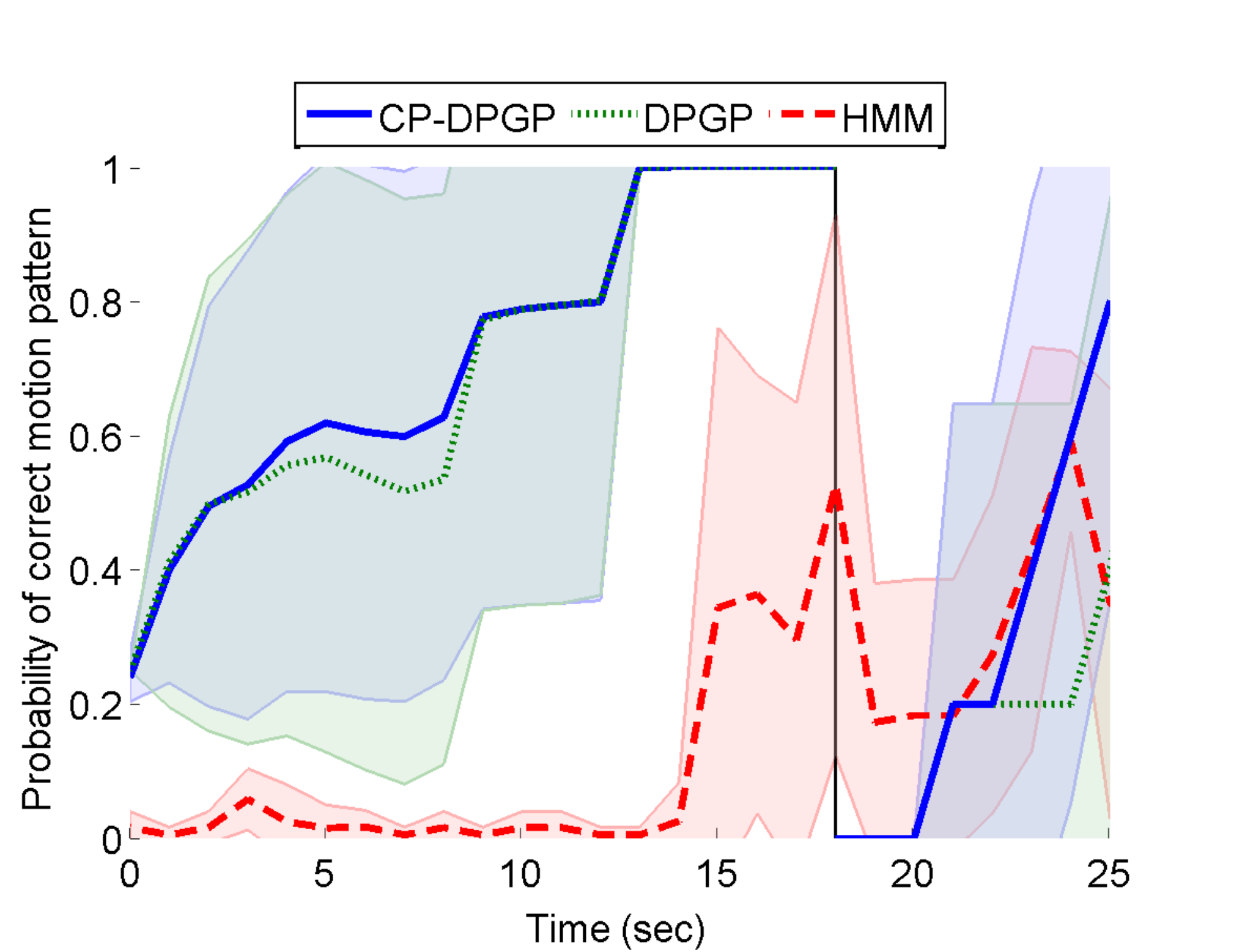}
 \label{fig:switch_intent}
 }
\subfigure[RMS predictive error]{
  \includegraphics[width=0.475\linewidth]{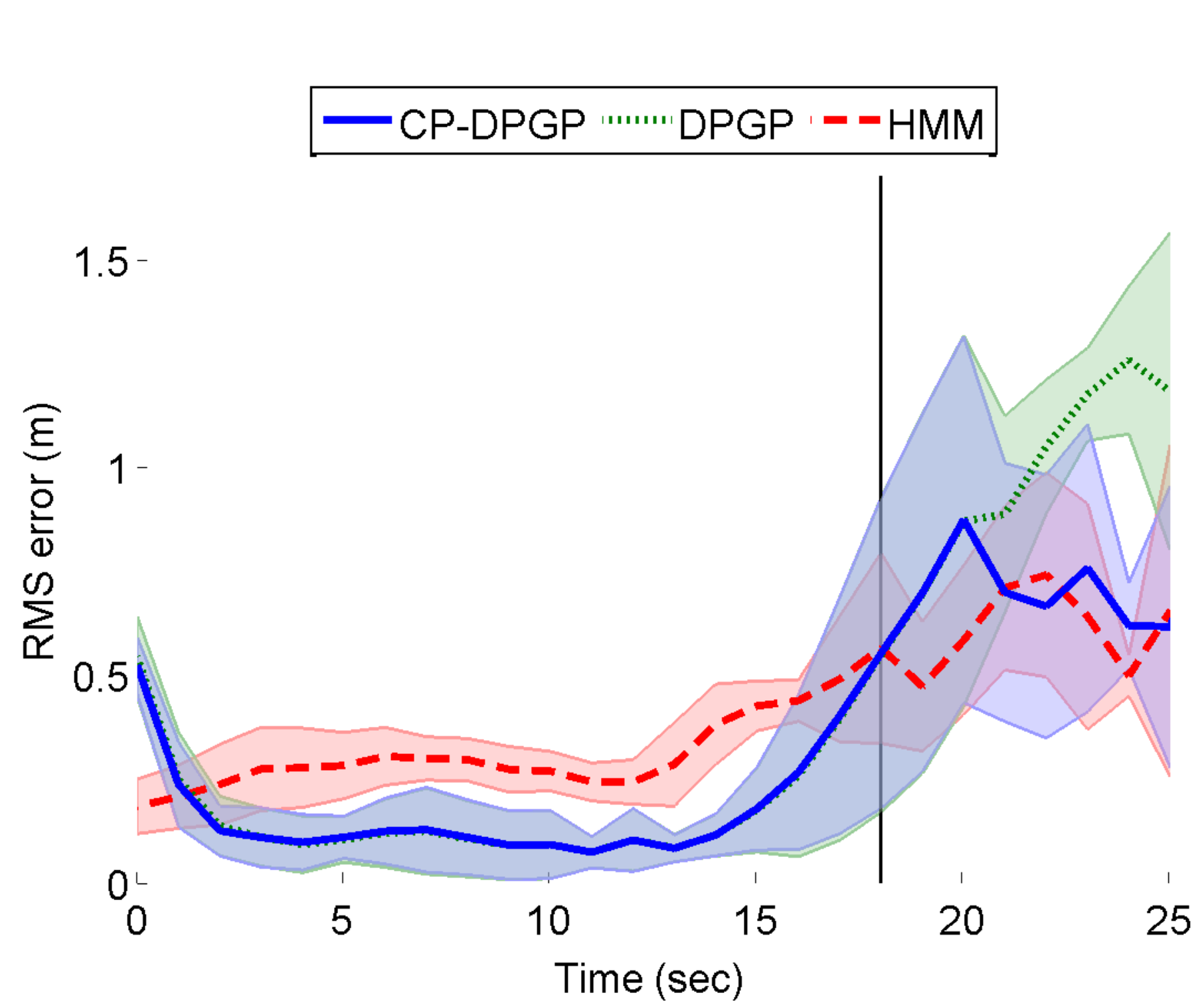}
 \label{fig:switch_rms}
 }
 \caption[]{Prediction accuracy of each algorithm for pedestrian crosswalk scenario, subject to pedestrian change in intentions at time = 18 s.}
 \label{fig:switch}
  \vspace{-2mm}
\end{figure}

Next, each algorithm is tested on five trajectories which demonstrate a change in pedestrian intentions, or a changepoint. In these trajectories, the pedestrian begins to traverse the crosswalk, but reverses direction after 18 seconds. Fig. \ref{fig:switch} shows the evolution of the correct likelihood and RMS error for each algorithm in this scenario, averaged across the trajectories. Both DPGP and Changepoint-DPGP converge on the correct behavior prior to the changepoint (Fig. \ref{fig:switch_intent}), while HMM performance is relatively unchanged compared to Fig. \ref{fig:baseline_rms}. As the change in pedestrian intention takes place, both GP-based algorithms initially drop to zero probability, as expected. However, DPGP accuracy remains poor beyond the changepoint, leading to the largest RMS errors of all algorithms (Fig. \ref{fig:switch_rms}). Because DPGP relies on the entire observation history, its predictions are slow to recognize the change, leading to worse performance. On the other hand, 
Changepoint-DPGP is able to selectively update the observation history considered in the likelihood computation given changes in intent, enabling it to achieve better accuracy than DPGP after the changepoint (Fig. \ref{fig:switch_intent}). As a result, Changepoint-DPGP yields the lowest average overall RMS error of all algorithms tested (Fig. \ref{fig:switch_rms}).

\begin{figure}[t]
\centering
\begin{minipage}{0.45\textwidth}
  \centering
  \includegraphics[width=\linewidth]{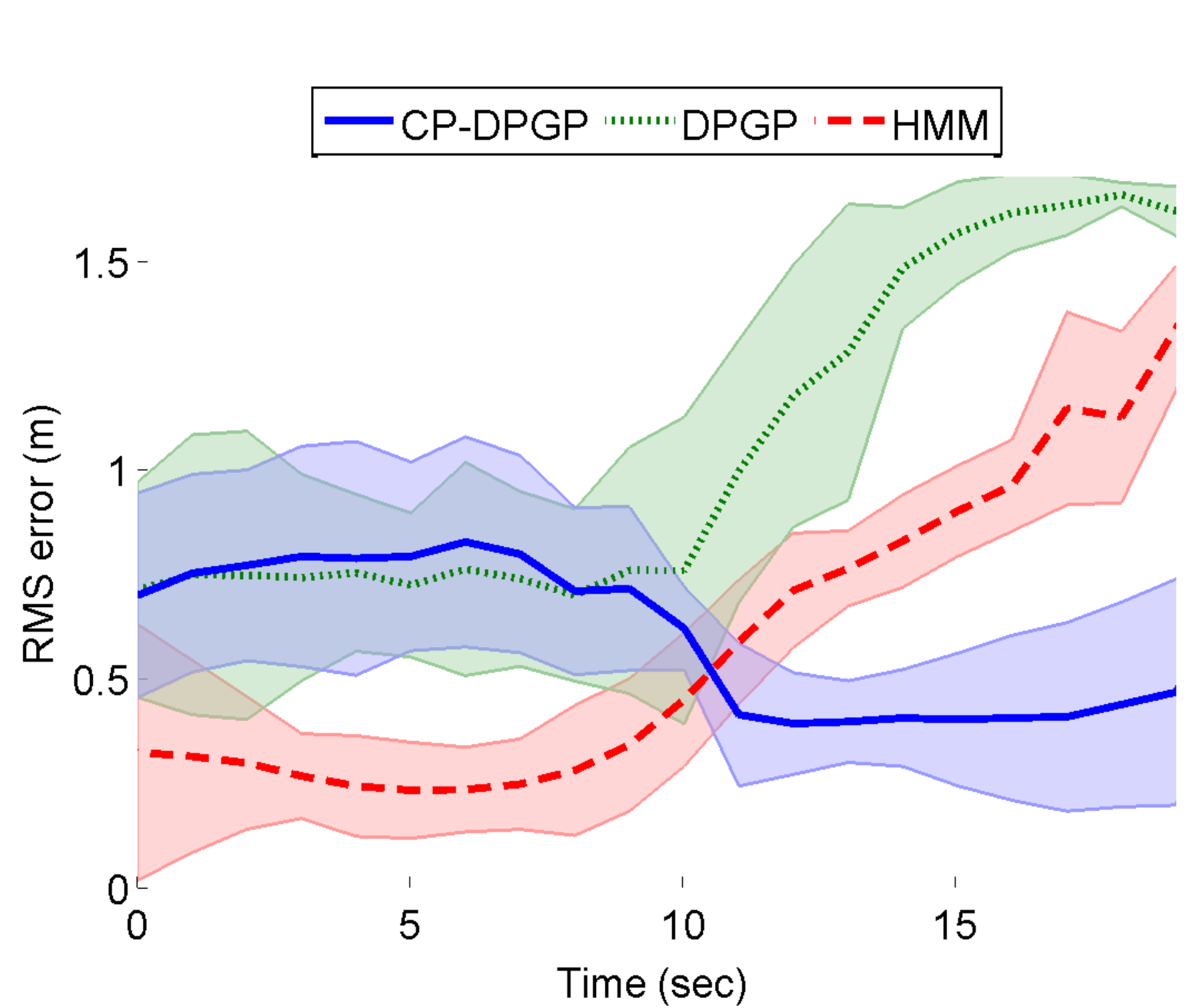}
  \caption[]{RMS error for pedestrian crosswalk scenario, subject to trajectories not observed in training data.}
  \label{fig:online_rms}
\end{minipage} \hspace{5mm}
\begin{minipage}{0.45\textwidth}
  \centering
  \includegraphics[width=\linewidth]{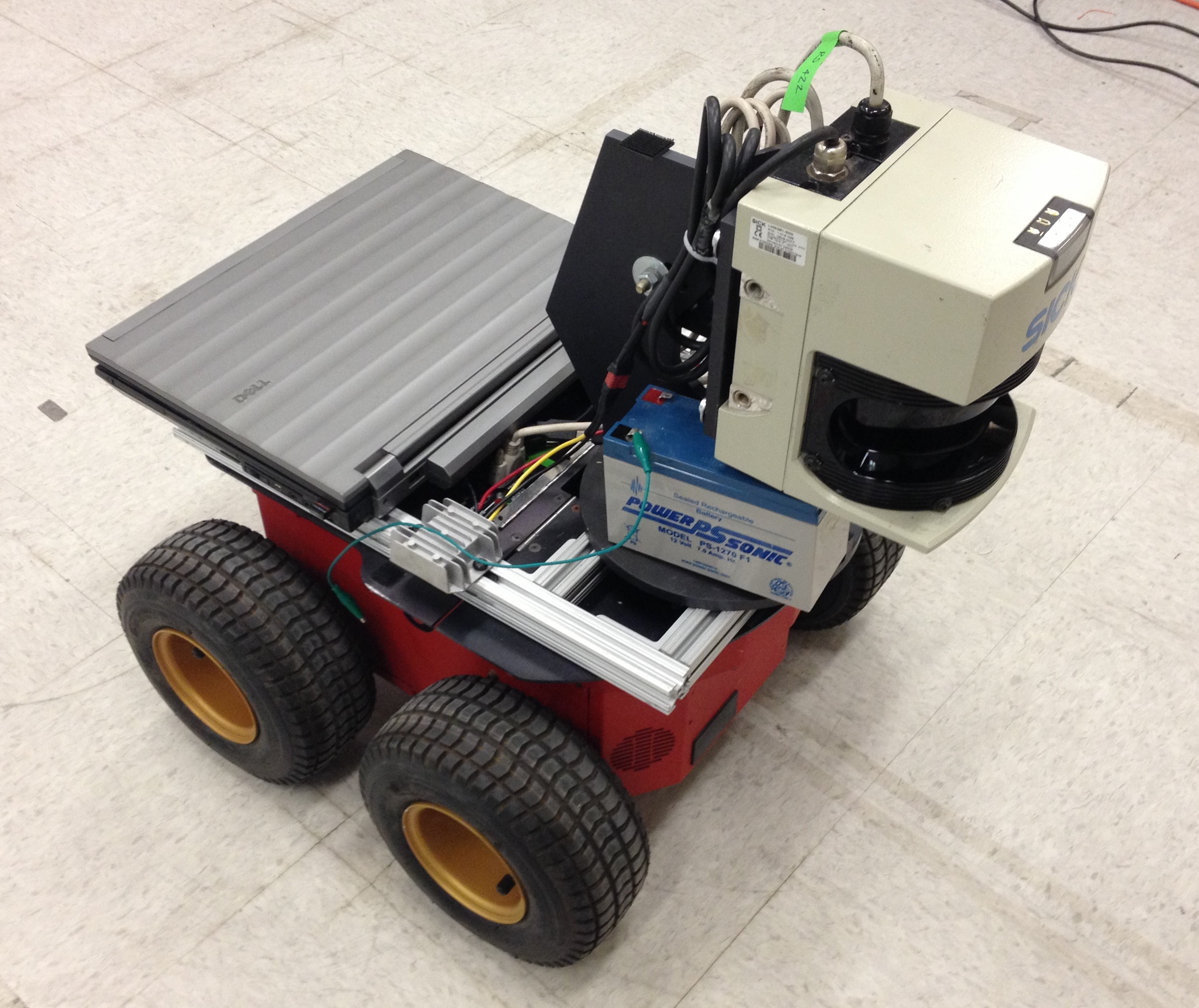}
  \caption[]{Rover used in closed-loop motion planning experiments.}
  \label{fig:rover}
\end{minipage}
\end{figure}

Changepoint-DPGP also demonstrates the best relative prediction accuracy when considering anomalous/new behavior patterns. In this scenario, algorithms are trained on only three of the four possible behaviors (red, blue, green in Fig. \ref{fig:crosswalk-velodata}), then tested on five trajectories from the fourth behavior (teal in Fig. \ref{fig:crosswalk-velodata}). The teal behavior deviates from the previously-observed red behavior approximately 9 seconds into the trajectory. Fig. \ref{fig:online_rms} shows the evolution of the RMS error for each algorithm in this scenario. At 9 seconds into the experiment, when the pedestrian behavior begins to deviate from anything observed in training data, the prediction of both HMM and DPGP begins to steadily increase. On the other hand, Changepoint-DPGP successfully identifies the new behavior and reclassifies subsequent trajectories. Thus it exhibits behavior similar to the baseline case, in which predictive error decreases as the 
probability of the correct motion pattern converges. Overall, Changepoint-DPGP predictive error is reduced by 62\% compared to DPGP. This demonstrates the strength of Change-DPGP in cases where the training data is not representative of the observed motion patterns, e.g. due to short periods of data collection.

\begin{figure*}[t]
 \centering
\includegraphics[width=0.33\linewidth]{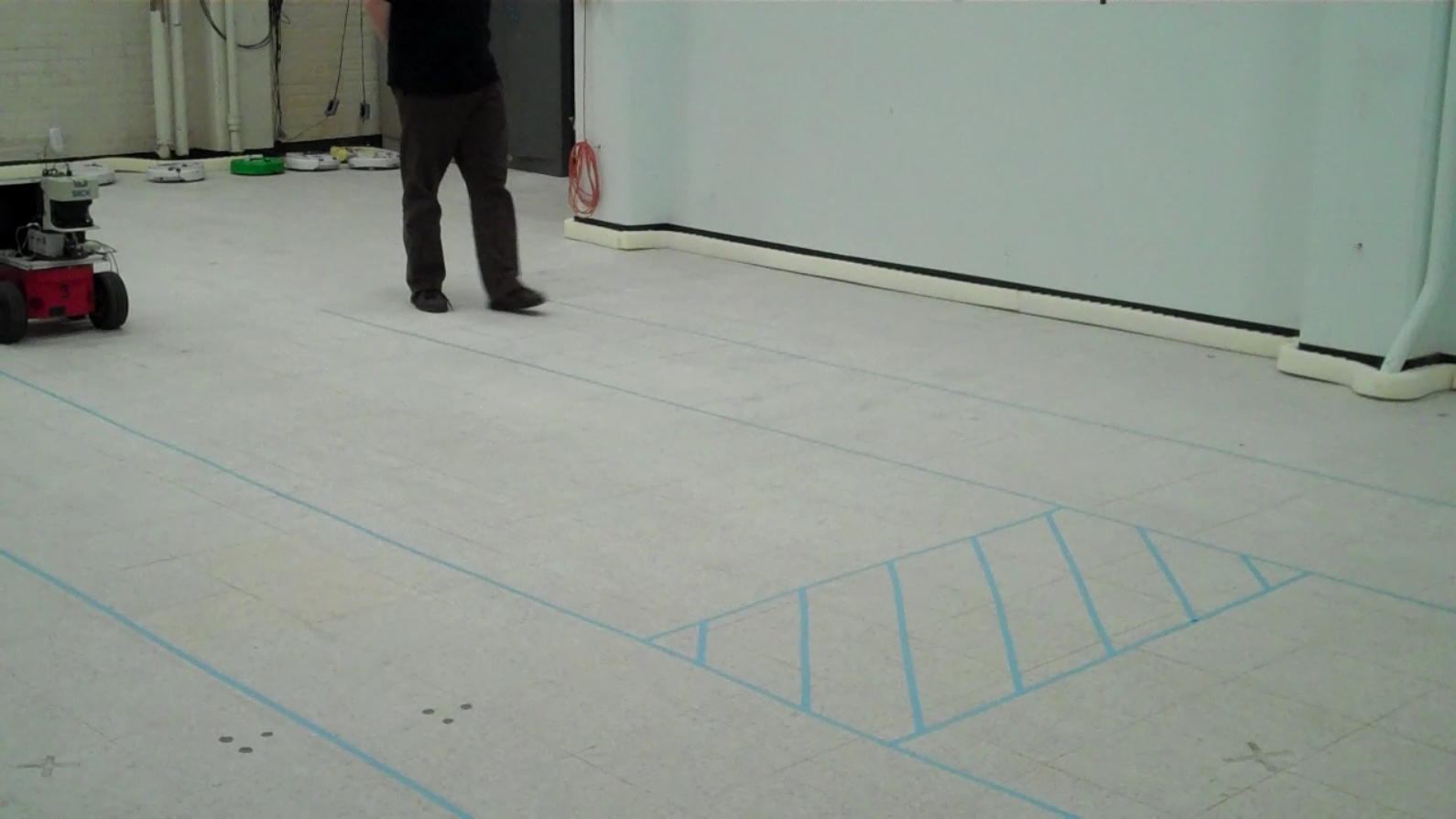}%
\includegraphics[width=0.33\linewidth]{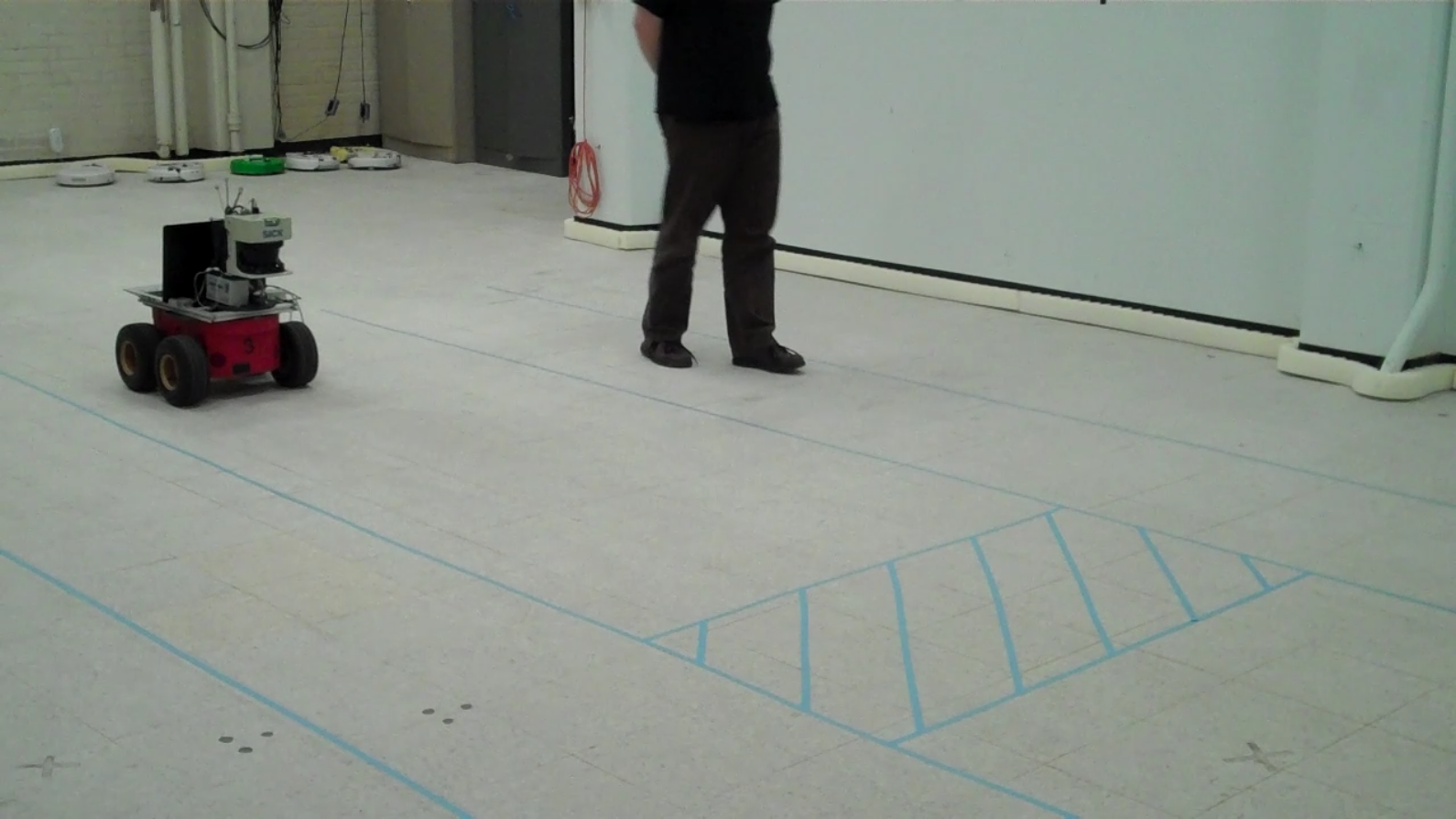}%
\includegraphics[width=0.33\linewidth]{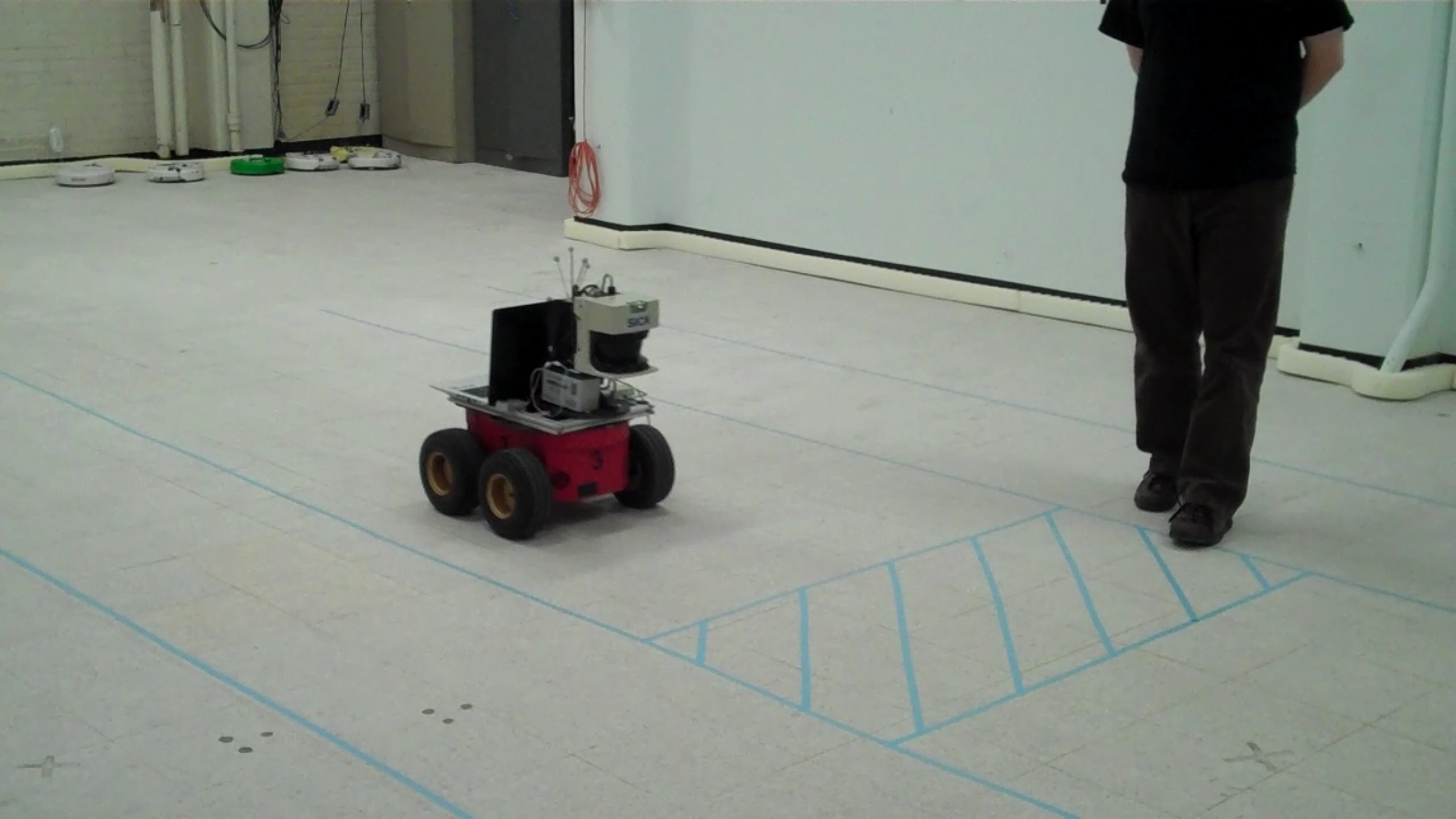}
\includegraphics[width=0.33\linewidth]{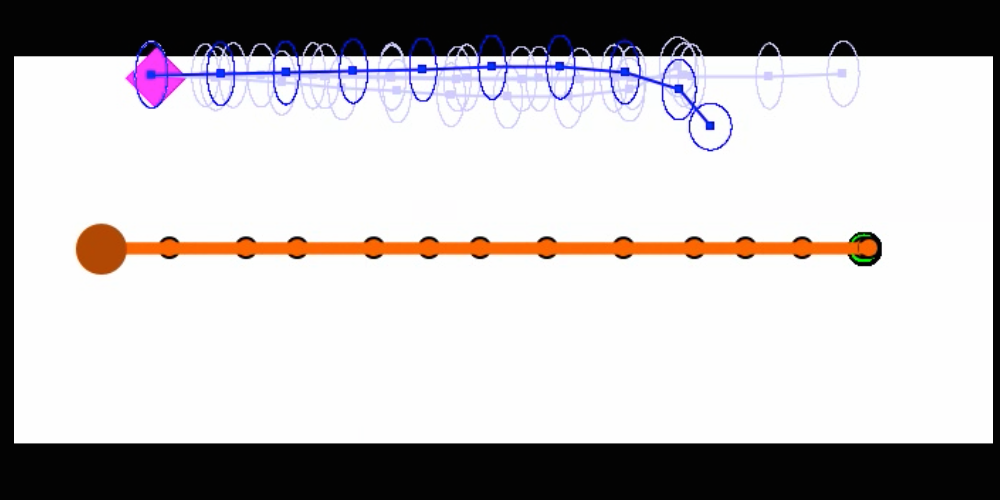}%
\includegraphics[width=0.33\linewidth]{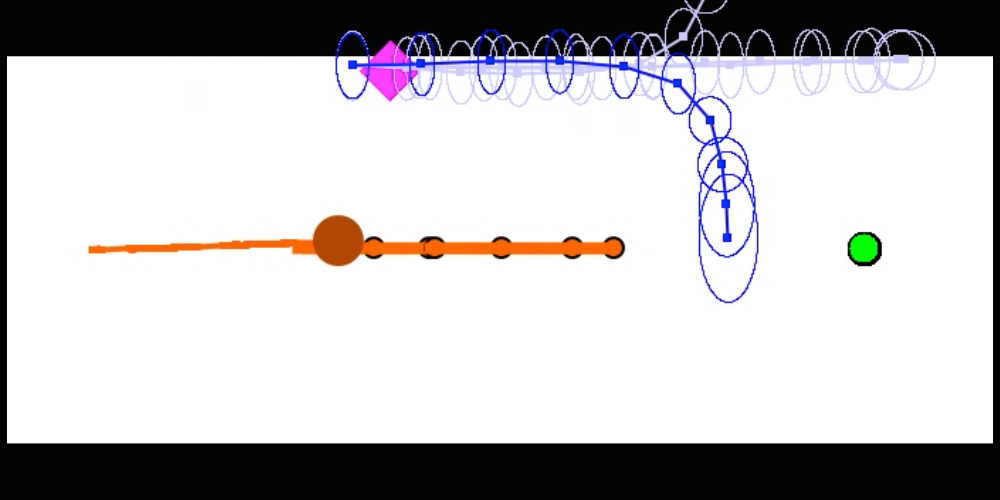}%
\includegraphics[width=0.33\linewidth]{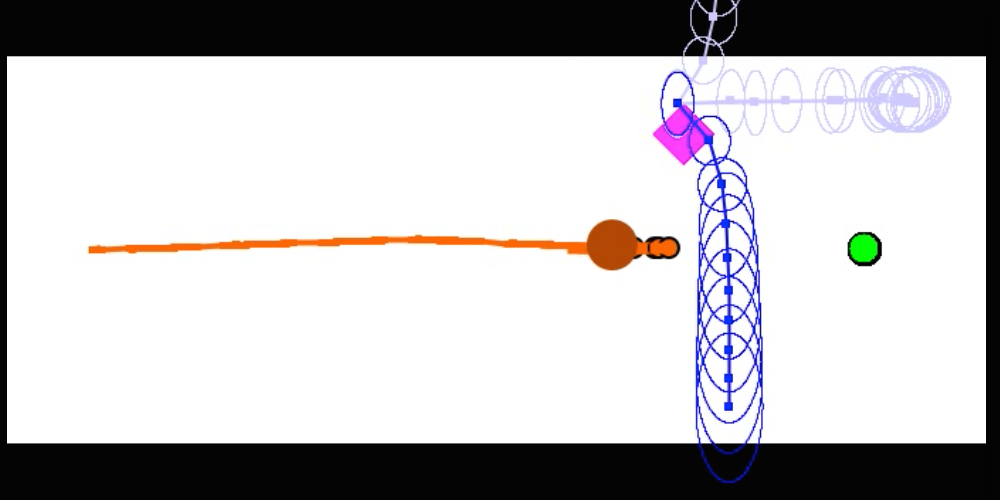}
\caption[]{Snapshots of a representative rover/pedestrian interaction, in which the autonomous rover safely avoids a pedestrian at the crosswalk.
The autonomous rover (brown) seeks to reach the goal location (green) while avoiding the pedestrian (magenta). The rover's current plan/history (orange) and predicted pedestrian behaviors (blue; darker shades indicate higher likelihoods) are shown.}
 \label{fig:crosswalk_live}
\end{figure*}

Finally, experiments have demonstrated that predictive results from the proposed Changepoint-DPGP enable an autonomous rover (Fig. \ref{fig:rover}) to safely avoid collision in closed-loop. Fig.~\ref{fig:crosswalk_live} gives snapshots of a representative interaction between a pedestrian and the autonomous rover. In this experiment, the rover localizes itself using motion-capture data, and identifies and localizes pedestrians exclusively from its onboard SICK lidar. The rover then uses the motion pattern predictions within a CC-RRT probabilistic motion planner (Sect. \ref{sec:ccrrt}) to generate safe trajectories. Initially, the planner generates a path directly to the goal, as the pedestrian is projected to remain on the sidewalk (Fig.~\ref{fig:crosswalk_live}, left). Once the predictions indicate that the pedestrian is likely to cross, the planner adjusts its plan to terminate prior to the crosswalk (Fig.~\ref{fig:crosswalk_live}, center). Once the pedestrian begins to cross (Fig.~\ref{fig:crosswalk_live}, 
right), the rover comes to a stop, waiting for the crosswalk to clear before safely proceeding to the goal.

\subsection{Multiple Dynamic Vehicles}
\begin{figure*}[p]
 \centering
\includegraphics[trim=0 0 0 0, clip, width=0.45\linewidth]{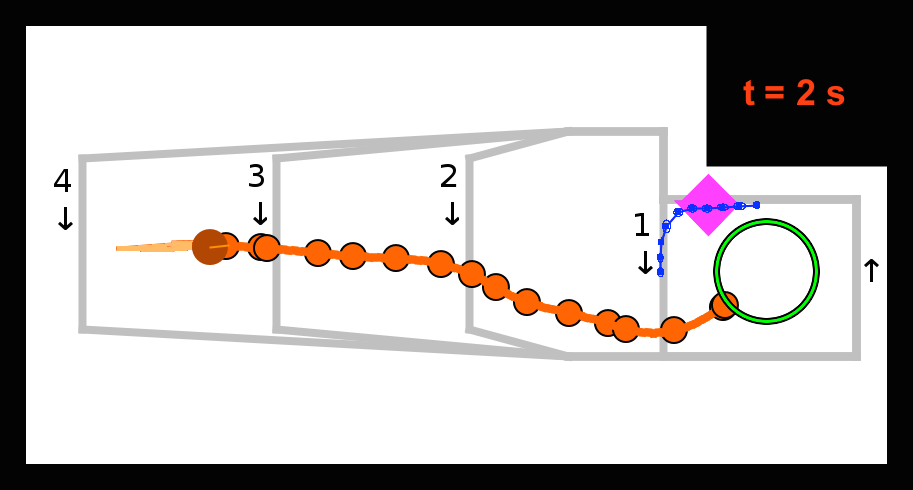}\hspace{10mm}
\vspace{-1mm}
\includegraphics[trim=0 0 0 0, clip, width=0.45\linewidth]{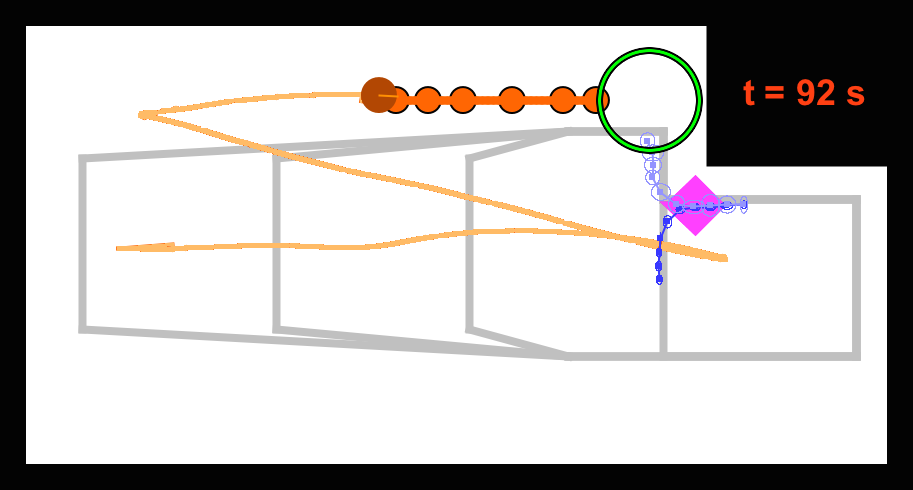}
\includegraphics[trim=0 0 0 0, clip, width=0.45\linewidth]{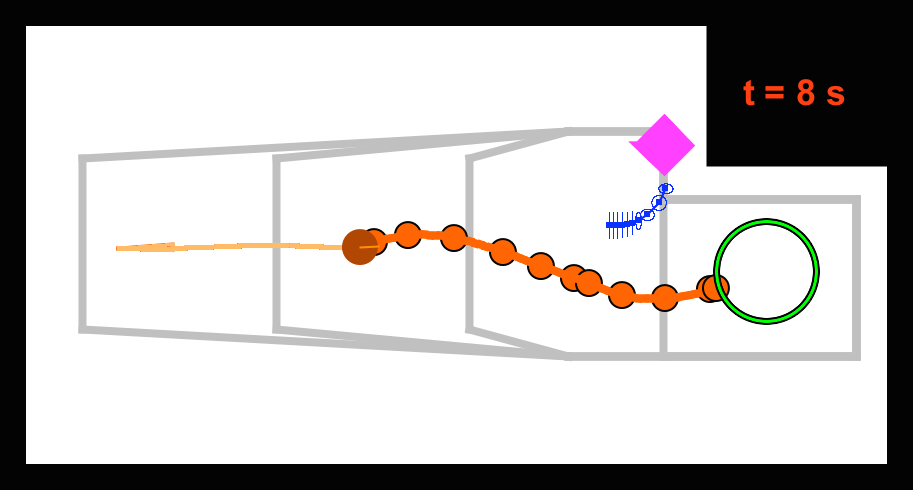}\hspace{10mm}
\vspace{-1mm}
\includegraphics[trim=0 0 0 0, clip, width=0.45\linewidth]{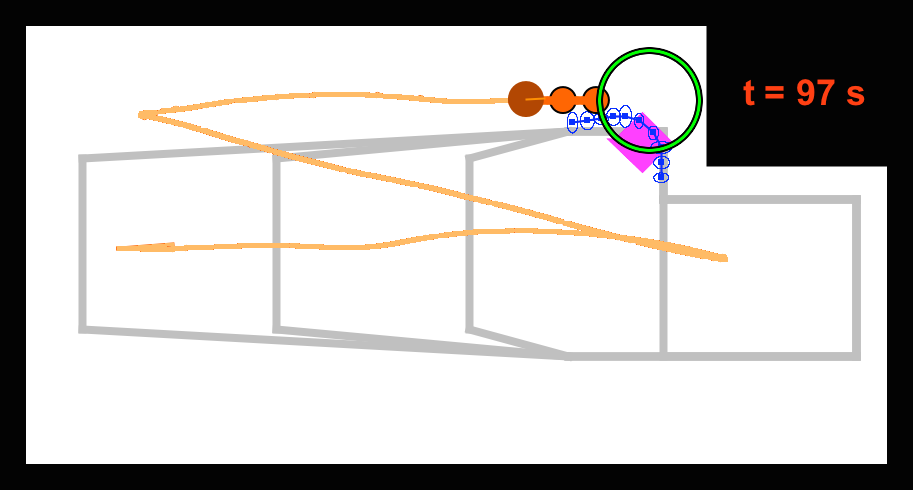}
\includegraphics[trim=0 0 0 0, clip, width=0.45\linewidth]{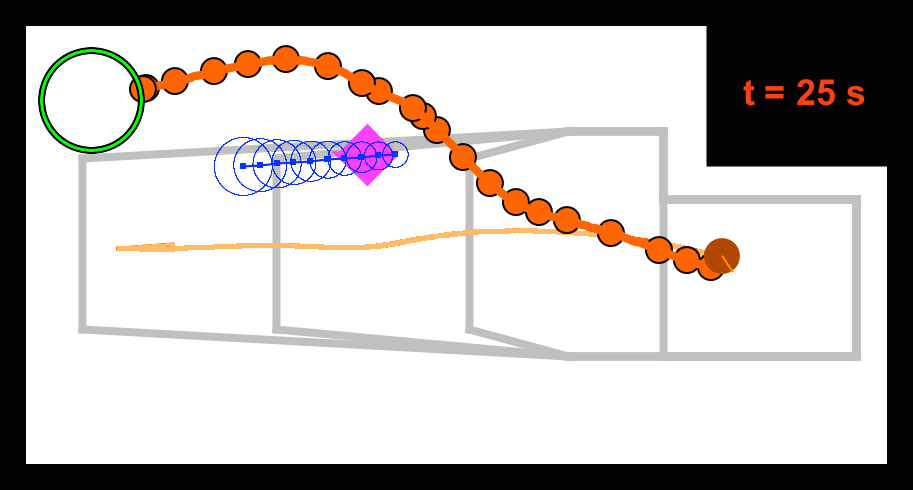}\hspace{10mm}
\vspace{-1mm}
\includegraphics[trim=0 0 0 0, clip, width=0.45\linewidth]{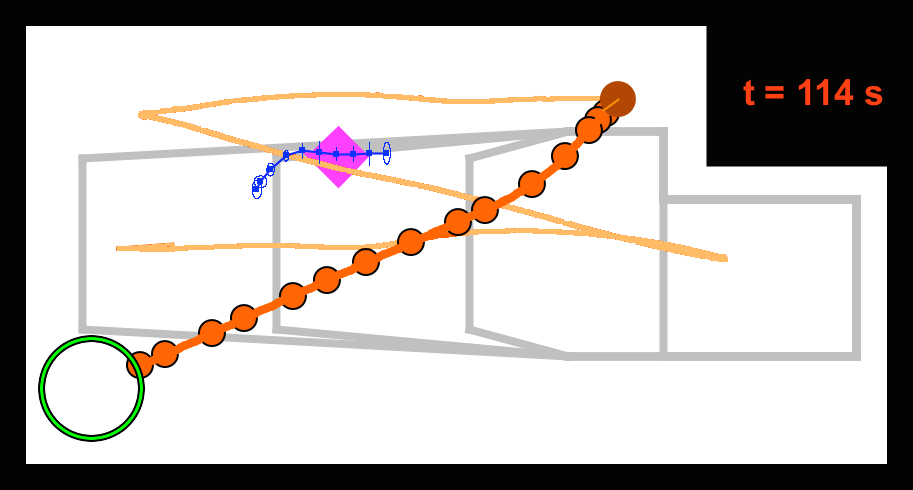}
\includegraphics[trim=0 0 0 0, clip, width=0.45\linewidth]{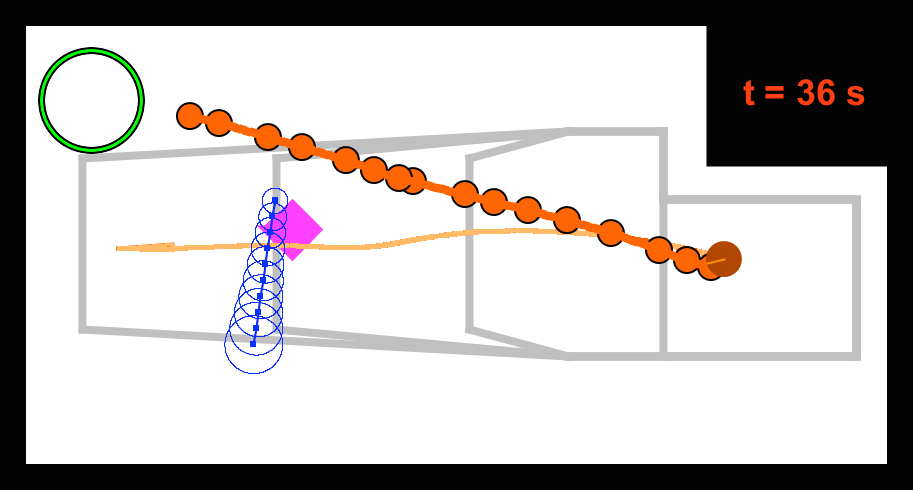}\hspace{10mm}
\vspace{-1mm}
\includegraphics[trim=0 0 0 0, clip, width=0.45\linewidth]{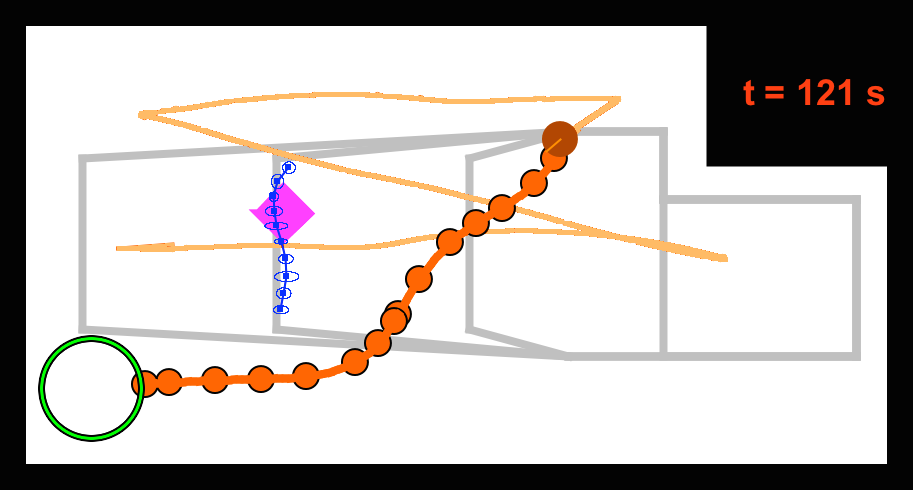}
\caption[]{Snapshots of a representative rover/robot interaction, in which the autonomous rover (brown) safely navigates to four goal regions (green ring) while avoiding a small dynamic robot (magenta). The rover's current plan (orange) and history (light orange) are shown, along with the predicted dynamic robot behaviors (blue; darker shades indicate higher likelihoods). Nominal dynamic robot behaviors are marked in gray; behaviors are numbered in first snapshot.}
 \label{fig:gpucc_live}
\end{figure*}

In this experiments, the autonomous rover must travel through a continuous sequence of waypoints while safely avoiding one or more dynamic robot obstacles (referred to below simply as ``robots''), using Changepoint-DPGP to predict robot motion and CC-RRT to generate safe trajectories. Each waypoint is represented as a goal region (Fig. \ref{fig:gpucc_live}) within the RAVEN motion-capture environment~\cite{How08_CSM}; the rover is provided with the next waypoint once the previous goal region is reached. The iRobot Create vehicle~\cite{irobotcreate} is used as the dynamic robot obstacle in these experiments. It continuously executes one of four cyclical, counter-clockwise motion patterns within the testbed (Fig. \ref{fig:gpucc_live}, first snapshot). State information for both the autonomous rover and iRobot is provided by the RAVEN motion-capture cameras.

Fig.~\ref{fig:gpucc_live} illustrates how Changepoint-DPGP behaves when a new behavior pattern is seen repeatedly. The rover has received training trajectories for behavior 1 offline, but a single dynamic robot actually executes behavior 3 online.  Initially, the only known behavior is behavior 1, so the autonomous rover is certain that the robot will pass between it and the goal and modifies its path accordingly. At 8 seconds, changepoint detection recognizes that the robot is executing a new behavior. Predictions are then generated assuming that the robot will continue at its current velocity with increased, linearly-scaling uncertainty. The planner modifies its planned paths to the goal to reflect this shift in perceived behavior.

After 85 seconds, the algorithm has learned the entire trajectory that it has just observed, as a new behavior. As the robot begins its second cycle, it still assigns the highest likelihood to behavior 1, based on the prior distribution of observed training and test trajectories still favoring behavior 1. However, behavior 3 is now included as an additional behavior prediction. By 97 seconds, the algorithm is fully confident that the robot is executing behavior 3, and shifts its likelihoods accordingly. The predictions for the new behavior accurately reflect the trajectory executed by the robot with reduced uncertainty. Based on this reduced uncertainty, the planner knows that the robot will turn before intersecting with the autonomous rover's planned path, and thus continues to execute that path. This scenario was executed for 2.5 minutes with no collisions.

\begin{figure*}[p]
 \centering
\includegraphics[trim=0 0 0 0, clip, width=0.45\linewidth]{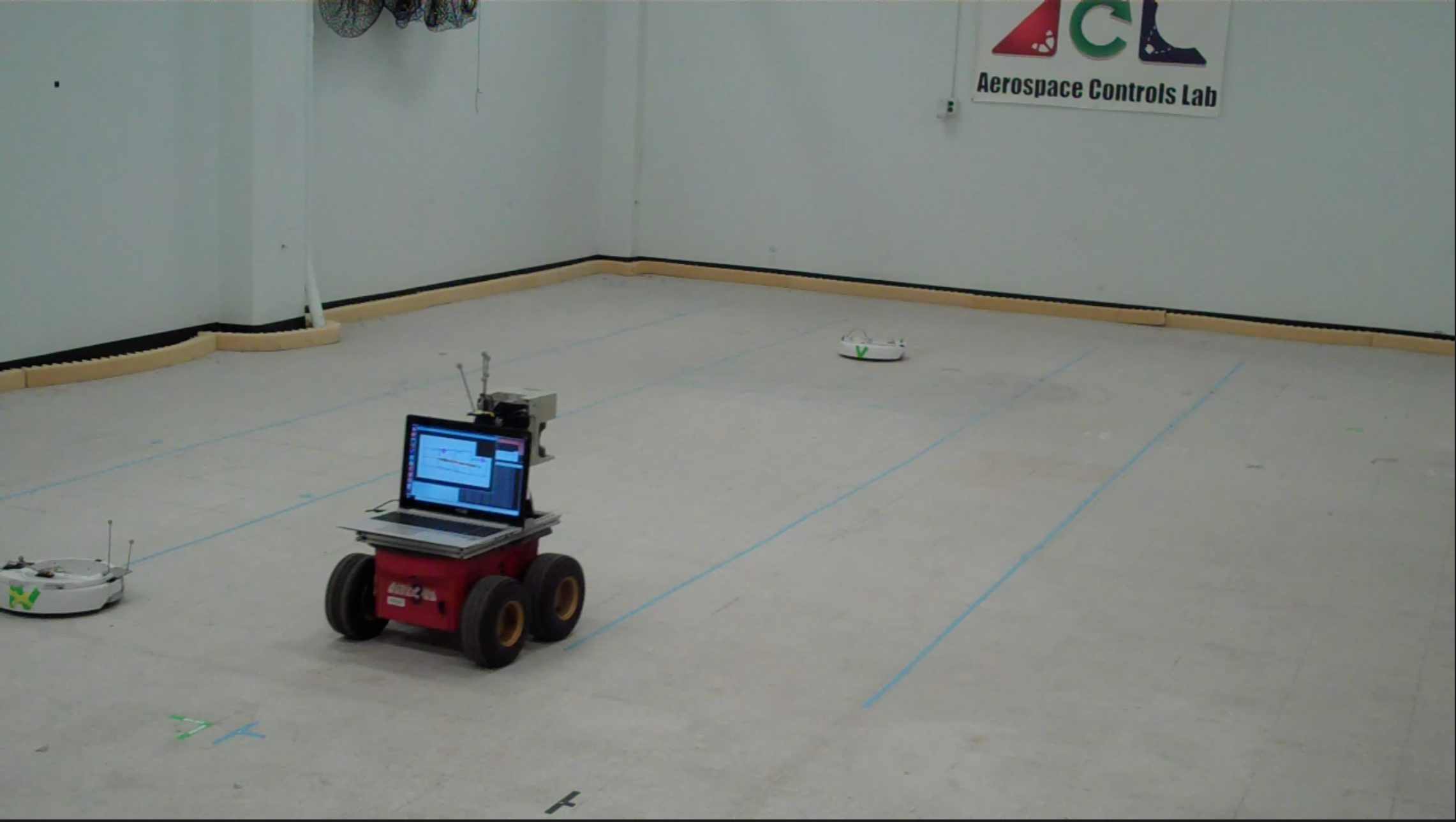}\hspace{10mm}
\vspace{-2mm}
\includegraphics[trim=0 0 0 0, clip, width=0.45\linewidth]{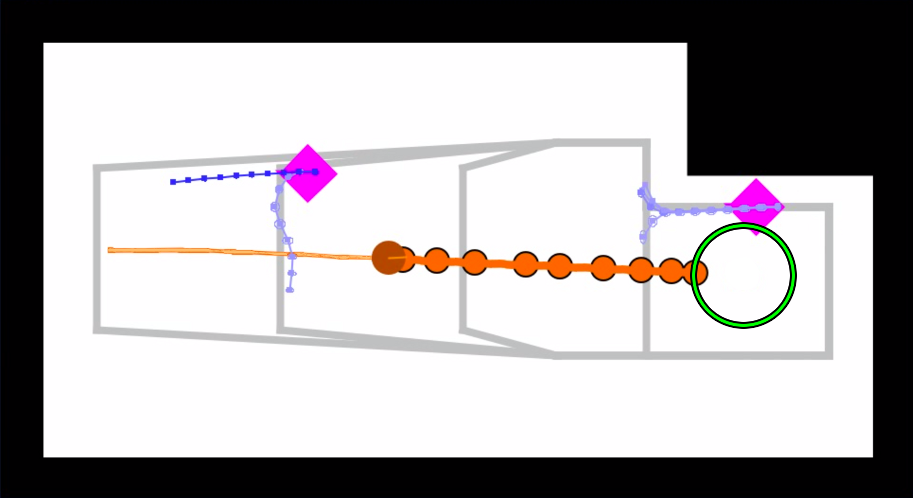}
\caption[]{Snapshot of a representative experiment consisting of the autonomous rover navigating to avoid two dynamic robots.}
 \label{fig:2obs}
\end{figure*}

Changepoint-DPGP has also been demonstrated to operate safely in real time with multiple dynamic obstacles; the final version of this paper will provide greater detail on this multi-robot examples. Fig.~\ref{fig:2obs} provides images from a representative baseline scenario, in which two dynamic robots are trained on and exhibit all four possible behaviors online. The scenario was executed for 2 minutes with no collisions. 

\section{Conclusions} \label{sec:conclusions}
This paper has developed a framework for long-term trajectory prediction and robust collision avoidance of pedestrians and other dynamic agents in real-time, even when these agents exhibit previously unobserved behaviors or changes in intent. A key contribution is the Changepoint-DPGP algorithm, which uses a non-Bayesian likelihood ratio test for efficient online classification and detection of changepoints. This algorithm is able to learn new behavior patterns online and quickly detect and react to changepoints, capabilities currently not present for other predictive algorithms that use GP motion models. As demonstrated in real-time simulation results, these capabilities significantly improve prediction accuracy relative to existing methods. Preliminary hardware results show that the framework can accurately predict motion patterns of dynamic agents from various sensor data and perform robust navigation.

\section*{Acknowledgments}

Research funded by Ford Motor Company (James McBride, Ford Project Manager).
%

\bibliographystyle{splncs_srt}
\bibliography{bibferguson,bibtex_Grande,ACL_all,ACL_bef2000,ACL_Publications}
\end{document}